# Grand Challenge: Real-time Destination and ETA Prediction for Maritime Traffic

Oleh Bodunov*, Florian Schmidt*, André Martin*, Andrey Brito#, Christof Fetzer*

*Technische Universität Dresden - Dresden, Germany
#Universidade Federal de Campina Grande - Campina Grande, Brazil
firstname.lastname@tu-dresden.de/andrey@computacao.ufcg.edu.br

## ABSTRACT

In this paper, we present our approach for solving the DEBS Grand Challenge 2018. The challenge asks to provide a prediction for (*i*) a destination and the (*ii*) arrival time of ships in a streaming-fashion using Geo-spatial data in the maritime context. Novel aspects of our approach include the use of ensemble learning based on Random Forest, Gradient Boosting Decision Trees (GBDT), XGBoost Trees and Extremely Randomized Trees (ERT) in order to provide a prediction for a destination while for the arrival time, we propose the use of Feed-forward Neural Networks. In our evaluation, we were able to achieve an accuracy of 97% for the port destination classification problem and 90% (in mins) for the ETA prediction.

## 1 INTRODUCTION

With the advent of autonomous driving, we are witnessing a massive growth of vehicle data being constantly collected from a variety of sensors, such as inductive loop detectors, GPS system, and mobile phones, promoting the development of data-driven intelligent transportation systems. Short-term vehicle prediction is currently one of the most dynamic research topics, aiming at estimating the traffic state in near future (within a few minutes) based on the historical data.

Although intelligent transportation systems are currently on vogue in car industry due to autonomous driving, prediction systems are also in high demand in other fields such as in the maritime context: The 8[th] Edition of the DEBS Grand Challenge [4] is taking this route as it focuses on the application of machine learning to spatio-temporal streaming data originating from vessels movements in the Mediterranean Sea. The goal of the challenge is to make the naval transportation industry more reliable by providing highly accurate predictions for vessels' destinations and arrival times improving the overall performance of the Supply Chain industry.

Despite being a well studied research field, ETA and destination prediction is not a trivial problem to solve. In fact, trajectory, speed, weather conditions and other external factors have often a strong influence on the actual trajectory and time needed to reach the final destination.

Previous research has shown that *machine learning* is a suitable approach when working with such kind of data as it allows the extraction of features that is often missed during manual analysis. Since machine learning comprises a large collection of methods, it is inevitable to perform a thorough analysis of the provided data set first in order to perform a proper partitioning and applying the right approaches to the given problem.

After analyzing the given data set, the two queries from the challenge, i.e., port and ETA prediction can be split into two separate problems: a *classification* and *regression* problem.

The prediction of the destination can be considered as a *classification problem*. Although regression models may seem appropriate at first, as they can capture intermediate states originating from the incoming Geo-spatial data, the task clearly asks for the prediction of high-level behavior which can only be carried out using classification. In contrast to the first query, the ETA prediction can be solved using regression.

In this paper, we describe our approach for predicting vessels' destination using a voted ensemble learning, while for the ETA prediction, we propose the use of Feed-forward Neural Networks.

## 2 BACKGROUND

Although trajectory and trip duration prediction is a widely known and well studied problem, existing solutions are often not applicable to the given use case as they are often tailored to their specific context in order to provide the highest accuracy.

In general, machine learning techniques for dealing with labeled data as provided by the challenge can be split into two categories: (i) classification and (ii) regression where each of them can be represented as one or another using appropriate feature engineering methods.

In the context of Geo-spatial data, many techniques have been explored previously such as Hidden Markov models [8], Kalman filter [3], support vector regression (SVR) [10], K-Nearest Neighbor (k-NN) [7] models, and data-driven models, such as artificial neural networks (ANNs) [9].

Although the previously mentioned methods are promising candidates for high prediction accuracy, ensemble learning as an alternative is one of the most popular and promising machine learning methods as it can improve significantly prediction quality by combining a large number of even weak base models.

In the context of transportation, tree-based ensemble methods that combine multiple simple decision trees have been successfully applied previously. Examples for such trees range from Random Forest, Gradient Boosting machine to Boosted Regression Trees.

## 3 APPROACH

In the following section, we will first provide details about our approach on how we reliable predict the destination port of a vessel using ensemble learning while in the second part, we provide an overview of our approach for the ETA prediction using Neural Networks.

### 3.1 Destination Port Prediction

Since vessel trips rarely follow a random trajectory, we can consider the prediction of a destination target port as a classification problem rather than predicting future coordinates for longitude and latitude using regression. In our evaluation and comparison of the two approaches, the first choice proved to be more accurate as it also reflects the behavior of the real-world travel patterns.

However, in order to capture the complex non-linearity of a trajectory in the context of vessel movements while still keeping the state of high-level behavior, we used an ensemble of methods based on trees:

We first extracted additional features from the given data set and compared them in order to identify the most important ones to be used in an ensemble of Decision Trees. We chose tree models for performance reasons and their ability to deal with unnormalized data as often experienced when working with real-world data.

While evaluating the different models, we noticed that Random Forest models provides the best runtime performance. Although Random Forest models do provide quick answers, they perform poorly with regards to prediction quality, especially when applying them to data sets comprising complex routes with many in-between stops. Hence, prediction quality for such models can only be improved by preprocessing the given data set such as separating each individual vessel trip.

We therefore experimented with alternative algorithms, however obtained only satisfying results when using the following ensemble method: We used a combination of Random Forest, Gradient Boosting Decision Trees (GBDT), XGBoost Trees and Extremely Randomized Trees (ERT) as they provided the best results even for complex trips with several in-between points that were not present in the set of destination ports.

We implemented the models using Scikit-learn [1], a Python-based machine learning framework. Prior to the training, we first mapped ports to numerical values which serves two purposes: First, the mapping can be used for an evaluation of additional features as well as for a reverse mapping used when processing the data in a streaming fashion. In order to find the optimal set of parameters for each model, we applied GridSearch.

During our evaluation, we observed that Random Forest Trees perform better when they were allowed to grow unbounded in size, while more complex algorithms such as GBDT and XGBoost perform better using smaller trees and did require additional parameter tuning.

For each of the models, we observed an accuracy of more than 92% (97% for Random Forest). In order to combine the classification results of the individual models, and to balance and overcome the weaknesses of one or another model, we used a Voting Classifier to obtain the final result from the ensemble. The final result of the ensemble is obtained through *hard* voting of the predictions obtained from the base models where *hard* simply refers to the fact that the predicted class label is selected by a majority rule voting.

We futhermore trained our ensemble using a transposed dataset that consists solely of numerical values and where all missing values were either dropped or substituted. Also, departures and arrival ports were transposed into numerical values using a lookup table.

The previously mentioned preprocessing step allows us to directly feed our tree-based models in a streaming fashion as required by the challenge. With regards to the number of features we used, we limited it to eight in total as it proved to provide already the best prediction accuracy. The features we used for classification task are as follows: ShipType, Speed, Lon, Lat, Course, Heading, DeparturePortName, ReportedDraught.

As an alternative, we also implemented an advanced deep learning model with Long Short-Term Memory (LSTM) cells that can be added to the top of the ensemble, however we discarded this model once an updated training dataset was released where each vessel trip was separately labeled.

Another reason for dropping LSTM model was its runtime overhead due to its complexity. Furthermore, in order to work properly, this model also requires a batch of incoming tuples for each particular vessel in order to perform a correct prediction of class labels. However, we left this model for future work as it can potentially achieve promising results for offline vessel prediction systems.

## 3.2 Arrival Time prediction

The second part of the challenge asked to provide a travel time/ETA prediction. As mentioned previously, this problem has been approached as a *regression* problem as follows:

For applications that must provide predictions in soft real-time and continuous manner as given by the challenge, travel time/ETA can be obtained by summarizing the section travel times at the current position where the overall travel time can be obtained once the trip has completed. However, since the velocity of a vessel is generally dependent on several factors, only time features can capture the exact behavior as they describe best the situation on each segment during the trip.

In order to overcome this problem, we applied *feature engineering* using the time-based data in order to extract the travel duration from each given timestamp to complete a particular trip. This DURATION feature (given in minutes) was attached to each row in the dataset and allows us to provide an ETA prediction using regression.

For the training phase, we used three additional features such as the destination port name and its longitude and latitude coordinates. In order to predict the remaining trip time, we utilized a FEED-FORWARD NEURAL NETWORK.

### 3.2.1 Neural Network Architecture.
Feed-forward neural networks — also called multilayer perceptron - can be visualized as a series of connected layers that form a network connecting an observation's feature values at one end, and the target value at the other end. It takes as input a set of fixed-size vectors and processes them through one or several hidden layers that compute higher level representations of the input. Finally the output layer returns the prediction for the corresponding inputs. For the given use case, the input layer receives initial numerical features such as coordinates, timestamp, course, heading, speed, and encoded values of departure and arrival ports from the dataset, and features that we found useful for regression task as described above.

We used standard hidden layers consisting of a matrix multiplication followed by a bias and a nonlinearity. The nonlinearity we chose is the LEAKY RECTIFIER LINEAR UNIT (LeakyReLU). We tested many different architectures and hyper-parameters.

Neural networks are typically created and initialized with small random values. We tested our model using different initializers, however the normal initializer [5] provided the best accuracy. We evaluated our deep learning models using KERAS and a TENSORFLOW backend. During the evaluation, we noticed that several of those capture similar results. The model that provided the best accuracy consists of a single input layer of 200 neurons and one hidden layer of the same size. As a loss function we used mean squared error. In order to minimize the loss, we furthermore used a RMSProp optimizer with learning rate $0.001$ without decay [6], and a batch size of 128. At each layer of the Neural Network we applied Dropout of 20% and trained our network using early stopping.

## 4 IMPLEMENTATION DETAILS

In this section, we will describe models we used in order to provide a prediction for the destination port as well as for the ETA. As mentioned previously, our model is based on ensemble learning where an ensemble of different models is used for short and long-term travel prediction. The first part of the algorithm predicts the destination port while the second part takes the prediction obtained from the ensemble in order to make its prediction for the time-delta, that describes the actual time in minutes for the ship to reach its destination port.

The final ETA is derived by adding the time-delta with timestamp from the incoming tuple. To assess the learning performance of the models, we randomly selected 70% of trips, withholding the remaining 30% as testing and validation sets (with a ratio of 15%). For prevent *data leakage* into the training set the whole dataset was sorted by different trips to ensure that ships that were present in training set were absent in other sets. The validation set was furthermore neither used during the training nor the testing phase but to validate final prediction accuracy of each model described in this paper.

*Feature Extraction*

In order to improve the accuracy of our approach, we drove a deep analysis of the data. The initially provided dataset contained several erroneous trips, as well as missing data in different fields. For example 60% of the data in the REPORTEDDRAUGHT field was not available. According to the description of the challenge, the minimum value for reported draught is 20, however, in order to decrease the bias of our model, we decided to fill the missing fields with a value of zero.

We also found that the initial dataset contained several incorrectly timestamped data records. Therefore, before training our models, we identified and removed all records with timestamps that were much higher than the arrival time or had time-flow problems such as time stamps from the past.

Moreover, all coordinate features were rounded to their $2^{nd}$ decimal which results in a precision for each coordinate that is sufficient to separate a village from another one on land. This step significantly decreased the variance of the spatial data.

In order to increase the performance of the ETA prediction, we used the following additional feature engineering techniques:

First, we split the trips by weekly, daily and hourly intervals, however, such a split did not reveal any useful information as all trips had an almost equal distribution over weeks and days etc. Furthermore, we omitted a deeper investigation with regards to monthly patterns as the dataset contained only data from a few months in 2015 where almost 80% of all trips appeared to happen in March. The most useful feature was the pre-computed time-delta for each ship from its current position to reach its final destination. This additional feature revealed to be useful and was included in the final model. Daily, weekly and monthly patterns proved to be helpful for training neural network (NN), as the architecture of a NN is good at extracting useful predicting power from a variety of features. Both approaches improved slightly the accuracy of our model, however, as a variety of additional features is required to computed *on-the-fly* in streaming systems, we decided to move on with fewer features for the sake of run-time performance.

As an alternative approach for the extraction of additional features, we used a MEAN-SHIFT clustering as described in [2]. We computed clusters for the whole area of where vessels were captured. This provided us with an encoded view of the areas in the Mediterranean sea where each longitude and latitude is mapped to a particular cluster. We obtained a set of 4817 clusters covering the whole area where each destination port (labeled or unlabeled) is mapped to a set of clusters. This approach allows to capture previously unseen ports and the resulting cluster information can be used to predict the next neighboring or destination clusters with a e.g. a LSTM model. However, using the labeled dataset, these features did not significantly improve the accuracy of any of the models.

*Model for Time Prediction*

In order to provide an accurate ETA prediction, we first analyzed the given data set, completed missing data items, encoded categorical variables and removed completely erroneous rows of data from the training set before initiating training of the model. Furthermore, we trained the NN models using a target variable time in minutes between the cut-off time point of our training snapshot and the timestamp associated to the last point of the trajectory. Before feeding our features into the NN, we normalized and re-scaled each feature individually such that it lies in an interval between zero and one. Using those preprocessing steps, we significantly improved the training time and convergence of our model. Moreover, extracting the duration for every timestamp allowed us to use the final model for the prediction of the arrival time not only from the past trips but also for trips in real-time.

## 4.1 Conclusion & Future Work

At the time of writing this paper, our model was trained on fixed dataset with around 300.000 tuples from over 300 ships. Using this training and validation set, we achieved an accuracy of 97% for port destination classification and 90% for ETA in minutes.

Our final prediction model for the two queries consists of an ensemble of machine learning models and a deep learning model.

This design allows the processing of streaming data where the destination port is predicted first from a set of given arrival ports, followed by the ETA.

In future, we plan to extend our LSTM approach for offline prediction. Many separate LSTM models also can be trained for each particular ship type in order to capture its unique behavior. The outputs of all these models can be fed to fully connected layers to predict n-timestamps in the future. This architecture is promising, however in order to be used with streaming systems additional work is needed.

In case the trip date contains several trips which was the case for the intially provided data set, a clustering approach can be used to capture previously unseen ports and improving accuracy.

Inspired by sequence-to-sequence models, we are also planning to improve our ETA prediction even further by predicting the next N-timestamps of the ship to gain insight not only of ETA but also on trajectory behavior for these next N-steps.

## 4.2 Acknowledgement

The research leading to these results has received funding from the European Community's Framework Program Horizon 2020 under grant agreement number 690111 (SecureCloud), 692178 (EBSIS) and by the German Excellence Initiative, Center for Advancing Electronics Dresden (cfAED), Resilience Path.